# Examining the Relationship between Scientific Publishing Activity and Hype-Driven Financial Bubbles: A Comparison of the Dot-Com and AI Eras

Aksheytha Chelikavada[1*], Casey C. Bennett[1]

[1]Department of Computing & Digital Media, DePaul University, Chicago, IL USA

*Corresponding Author: Aksheytha Chelikavada, achelika@depaul.edu

## Abstract

Financial bubbles often arrive without much warning, but create long-lasting economic effects. For example, during the dot-com bubble, innovative technologies created market disruptions through excitement for a promised bright future. Such technologies originated from research where scientists had developed them for years prior to their entry into the markets. That raises a question on the possibility of analyzing scientific publishing data (e.g. citation networks) leading up to a bubble for signals that may forecast the rise and fall of similar future bubbles. To that end, we utilized temporal SNAs to detect possible relationships between the publication citation networks of scientists and financial market data during two modern eras of rapidly shifting technology: 1) dot-com era from 1994 to 2001 and 2) AI era from 2017 to 2024. Results showed that the patterns from the dot-com era (which did end in a bubble) did not definitively predict the rise and fall of an AI bubble. While yearly citation networks reflected possible changes in publishing behavior of scientists between the two eras, there was a subset of AI era scientists whose publication influence patterns mirrored those during the dot-com era. Upon further analysis using multiple analysis techniques (LSTM, KNN, AR X/GARCH), the data seems to suggest two possibilities for the AI era: unprecedented form of financial bubble unseen or that no bubble exists. In conclusion, our findings imply that the patterns present in the dot-com era do not effectively translate in such a manner to apply them to the AI market.

# 1. Introduction

## 1.1 Overview

Around the clock, vast flows of capital silently move through the world's financial platforms to fuel innovations. These flows matter because they have the power to change and reshape economies by impacting the fate of various industries [70]. For instance, they can cause certain industries to experience a boom over a short period of time where start-ups grow quickly into billion-dollar companies, yet however they can also trigger the collapse of industries if underlying fundamentals are weak [72].At the heart of these shifts is the stock market. Here, the ownership of assets (e.g. stakes in companies) is exchanged through the buying and selling of shares. But the stock market is unique, compared to other financial systems, because it allows public individuals (e.g. retail investors) and large institutions to compete with one another, to not only make deals on the future of the market to make profit, but to also influence the ebb and flow of the value of companies through their decisions [21]. Each of those decisions is essentially a judgement (i.e. "estimate") of a company's relevance and role in the market. When all of these judgements are combined, they in turn go on to create powerful signals to other market actors regarding the company's direction and its performance in the economy [61]. This sort of accessibility allows the market to reflect the overall sentiment and strategic behavior of a diverse array of investors through every single buy or sell action. Therefore, the stock market simply does not just reflect the news, but rather it absorbs the emotions and gut-feelings of investors to showcase a real-time picture of what people think of the economy [47].

However, when emotions take precedence over thorough fundamental analysis and evaluation, markets start to behave unrealistically [1]. In the modern era, that exuberant behavior is often linked to technological advances, and thus by proxy to its underlying scientific progress. This usually occurs when there is delusion over inflated values of assets and that they will keep rising while ignoring fundamental warning signs like increasing debt burden and unrealistic business models/plans [38]. As a result, stock prices no longer reflect the realistic value of industries. Rather, prices of assets start to uncontrollably rise and create false confidence which can go to inflate the prices more, creating a vicious feedback cycle. This type of cycle that is fueled by irrational speculations and exuberant excitement is called a financial bubble [3]. These bubbles will burst at some point in time and often will do so very abruptly [55]. **That raises an important scientific question as to whether publication patterns of scientists working on innovative technologies prior to past bubbles could be used to predict forecast potential market disruptions in the future before they happen?**

A popular example of the phenomenon described above is the dot-com bubble which took place in the 1990's and early 2000's. During this time, investors became obsessed with the future promises of web technology such as web applications, internet services, network infrastructures, and online marketplaces, leading to a growing hype that outpaced actual technical developments [6]. At the time, most of those innovative technologies had roots in institutional research where scientists worked on them for many years prior within academic institutions or research institutions. So, in a way, we could hypothesize that the signs of that impending bubble should already have been visible in research publication patterns in the years leading up to it (e.g. publication rates, size of citation networks, etc.). In other words, there could have been patterns in the publications about web technology that might have hinted at a potential change in the markets.

Perhaps there are similarities across bubbles, for instance. However, there could be a more complex, non-linear relationship between markets and technical research. Due to progress in academia being sometimes slow to transition into the development of new technologies and products in industry, it often takes a very long time (sometimes years or decades) for them to become marketable whereas financial markets are fast-paced and prone to sudden changes in government policies and/or geopolitical tensions. We intend to explore these various possible hypothetical relationships in this paper through a data-driven approach, using multiple analyses methods.

### 1.2 Background Research on Tech Market Bubbles

The phenomenon of financial bubbles has been much studied because they significantly impact the stability of the economy, which then leads to severe consequences for many people as it affects their financial security. This includes the loss of jobs, diminishing savings, and restricted access to credit, all of which can impact people's ability to support their basic needs and plans for a secure, stable future [9]

Previous research has found that during the dot-com bubble, analysts were overly optimistic and were driving the stock prices of technological companies to an all-time high, which the federal chairman at that time, Alan Greenspan, described as "irrational exuberance" [52]. By using statistical methods such as the adjusted R-square from regression models, prior research has shown that during the bubble fundamental analysis did not explain much of the *value relevance* for many dot-com tech companies, which refers to the use of traditional financial indicators such as earnings and assets to reflect a company's market value when compared to the estimates of analysts. However, after the bubble burst, the adjusted R-square increased. This suggests that the dot-com bubble was driven by irrational behavior based on technological hype and a disregard for traditional fundamental analysis [38].

The irrational behavior that was present during the dot-com bubble was very much driven and encouraged by media people, analysts, technology leaders, investors, and even some scientists. They were spreading narratives about a "new economy" being ushered in and justified the unusually high valuations of companies despite no profits or earnings. Therefore, many people went along with the dominant narrative and refrained from using data-backed findings as they were seen as outdated methods that only worked for "older" economies [22].

Such detachment from fundamentals could be witnessed through the underpricing of IPOs. This is when a company's initial public offering (IPO) price of shares is set by investment banks to be lower than the price they trade on when they are publicly offered so that large institutional investors immediately gain profit (retail investors, not so much). This was a very significant issue in technological companies during the bubble in the late 1990's as underwriters would often favor certain clients and create incentives by severely underpricing shares. In turn, IPOs would attract attention and create media hype for companies. However, this is concerning because when stock prices inflate excessively with no support, the prices will crash down when investors cash out [30]. Financial bubbles are inevitable, but the impacts of having a bubble burst can increase the likelihood for other bubbles to develop. When a bubble bursts, banks will usually create incentives like cutting interest rates to support the damaged economy. Then, there is always a search for the other big investment opportunity. Hence, perfect conditions for the next bubble with investors aggressively looking for returns to make up for their losses [55].

Although the impact of scientific behaviors on the dot-com bubble has not been explored much yet, there have been recent technological breakthroughs that have created sudden changes in the stock market. An example of this would be the release of DeepSeek, which is an open-source AI model from China. The entry of this innovation destabilized the stock market and triggered a $ 1 trillion sell off in US and European tech-based stocks [14]. So, perhaps there could have been signals of this potential disruption in the behavior of scientific communities.

Today, with the boom and explosion of artificial intelligence (AI), there is a very real possibility that we are leading to (or are already) in an AI bubble. We are seeing many similar patterns from the dot-com bubble in terms of media hype, speculations, and extremely high valuations for some companies despite minimal or no profits [19]. Other research has found hints of speculative bubbles present in the Nasdaq stock market from January 2017 to January 2025 as AI continues to grow at a rapid pace [8]. However, it is difficult to actually confirm one is in a bubble during the middle of the bubble, which suggests a more sophisticated analysis approach is needed.

### 1.3 Research Aims

There are some financial markets that are known indicators for identifying bubbles, but those often only start to flash warning signs when it is too late. Another approach would be to utilize publication patterns of scientific innovation in the peer-reviewed literature (e.g. citation networks) as an early indicator, but that requires a method that can reliably derive those indicators from such data. **We propose here using social network analysis (SNA) to help to detect patterns and signals of bubbles within the scientific ecosystem**. SNAs are known for taking in large-scale, complex data to detect communities and analyze their evolution over time. By doing so, one can understand behaviors and hidden relationships that would have been difficult to detect without SNA [12]. SNAs also have the ability to be analyzed incorporating the dimension of time. These temporal SNAs allow the ability to study the progression of relationships and behaviors within the network over time as well as the changes in the social network structure itself [60]. This approach was especially helpful during the COVID-19 pandemic, during which there was a significant shift in behavior and sentiment towards the disease across human society. Hence, researchers applied SNA to X (formerly known as Twitter) data to analyze the shift of discussion topics, perspectives, and emotion sentiment through tweets and retweet across networks of user [25]. It is possible that SNA could work similarly for analyzing bubbles via scientific publication patterns and citation networks.

New innovations create market disruptions through excitement when they arrive with new promises and brighter futures. We saw that in the dot-com bubble with the arrival of breakthrough web technology [54]. Although such technology really did help evolve human society in significant ways, it did so in a painful manner with overvaluation and excessive excitement causing a devastating market collapse [24]. This indeed left a deep mark in the financial markets and the prospects of younger generations at the time. **That raises the question of whether by analyzing data leading up to the dot-com bubble and its burst, we could predict a similar AI bubble in the current era?** Doing so could perhaps help prevent or reduce the effects of such a volatile transformation today due to any such AI bubble. With many new innovations being developed in the scientific research space of academic institutions steadily bleeding out into commercial markets, it is possible that innovations can be tracked and anticipated from scientific research publication patterns, and likewise those patterns may foreshadow impending changes. If so, then policy makers could anticipate future changes in the market that could signal the existence of a bubble as well as the bubble bursts. This is especially important today, as AI is one of the fastest emerging technologies and is growing and changing at an unprecedented rate [63].

To put it more precisely: by analyzing scientific publication patterns and financial data from the dot-com era from 1994 to 2001 (including pre-bubble, bubble build-up, and post-bubble burst), we aim here to test whether it is possible to detect/predict similar patterns and signals in the stock market indices and research activity from 2017 to 2024 that can help forecast a potential AI bubble formation and burst***. We take the previous sentence as the central hypothesis of this paper.***

To test that hypothesis, we first gathered publication data from an open-source scientific publication database, OpenAlex. For the historical stock market indices, we pulled monthly market data for CBOE Volatility, S&P 500, NASDAQ Composite, and NYSE Composite from Investing.com during those time periods. Lastly, we used an online repository of IPO and venture capital data from Professor Jay Ritter at the University of Florida (https://site.warrington.ufl.edu/ritter/ipo-data/). That repository covers monthly number of IPOs and the average first-day return upon public release, both for traditional IPOs as well as venture capital backed IPOs. For the scientific publication data, Leiden network characteristics were utilized to measure collaborations and sub-domains within scientific communities. Afterwards, GCN (graph convolutional networks) were used to compare research activity levels during the rise of AI to that seen in the dot-com era. Then, we modeled AI market using Long-short term memory (LSTM) networks and K-nearest neighbor (KNN) classifiers to identify whether the research activity during the AI period followed the same pattern as seen in the market during the dot-com era.

We note that we decided to use the dot-com boom to predict any AI-era bubbles because both eras share several similarities. They both deal with extremely transformative technology that is/was filled with enormous promise, enticing investors to flock to investing in companies that are working with AI or dot-com technologies without solid business models or generated profits [19]. Additionally, the dot-com boom witnessed a surge in publications, IPOs, and venture capital funding in the years leading up to the bubble, which at least on the surface seems to resemble the field of AI today.

# 2. Methodology

## 2.1 Dataset Collection

### 2.1.1 Research Activity Data Collection

In order to begin our study on the research activity during the two technological booms (dotcom and AI), we identified timeframes for each one. We chose influential breakthroughs in scientific research prior to the visible indicators in financial indices for the dot-com bubble to establish the influence research has on financial market behaviors. For the dot-com era, that period was 1994 to 2001, as 1994 began a transformative time in digital technology as publications such as The World Wide Web by Berners-Lee et al. (1994) and NCSA Mosaic: A Global Hypermedia System by Andreessen et al. (1994) laid the foundation for many impending web technologies. The year 2001 was the burst of the dot-com bubble, as indicated in the literature [38]. Similarly, we chose a 7-year period for AI boom to match the same length, ending in 2024 (the time of data collection for this study). However, we also note that 2017 marked a time of revolutionary ideas and processes in AI that sparked interest of academia and investors [46], including seminal papers in 2017 such as "Attention is all You Need" that introduced transformer models and helped spark the rise of generative AI [66].

To analyze how influential research spreads and shapes the behavior of the scientific community, we selected highly influential papers based on citation counts from the OpenAlex publication database as the initial parent "root nodes" for our network and utilized snowball sampling [53]. That was done separately for the dotcom boom and the AI boom periods. To do so, we first selected the influential paper/s and then collected a maximum of 200 of papers (maximum limit placed by OpenAlex) that cited each parent paper. Though the max citation limit in OpenAlex was a limitation to this study, we note that previous research in Nature has shown that most published papers (~75%) have less than 10 citations and only 1-2% even reach 100 citations [65], so the limit likely had minimal impact. After collecting the citing papers of the parent node, we recursively collected the citations of the citing papers (i.e. citations of the citations of the citations etc.), so that we could gather the flow of innovation "impact" over time as measured by subsequent research activity [60].

To gather all the citation data, we scraped publication data from OpenAlex. By utilizing its API through Python, we collected the following information for each paper: title, publication year, DOI, OpenAlexID (unique identification code for each paper on OpenAlex), parentID (OpenAlexID of the paper the current paper cites), authors, affiliations of the authors, and affiliated countries of the authors. For the parent root node papers, the ParentID section was left empty. This process was conducted twice to create two datasets: one for dot-com era and one for AI era.

### 2.1.2 Stock Market Indices and IPO Data Collection

To identify if there is a relationship between research activity and economic behavior, we collected several historical stock market indices and IPO data.

Stock market indices are useful because they condense the performance of thousands of stocks into a few numbers. This then makes it easier to track behaviors of different market segments such as tech-heavy or blue-chip stocks under varying market conditions. The stock market indices used here were CBOE Volatility (VIX), S&P 500 (SPX), NASDAQ Composite (IXIC), and NYSE Composite (NYA), which were recorded monthly for each era and include date, closing index price, opening index price, highest index price, lowest index price, the volume of stocks traded, and the change percent, which is the percentage change from the previous month's closing price. The CBOE Volatility Index (VIX) was chosen because it reflects investor sentiment, such as uncertainty and/or fear through S&P 500 options. The S&P 500 Index (SPX) was chosen because by design it captures a broader reflection of US markets and serves as a bellwether for global markets. The NASDAQ Composite Index (IXIC) was chosen because it is the largest stock exchange by volume being traded and contains many technology-based companies, and it was heavily impacted during the dot-com. The NYSE Composite (NYA) was chosen because, in contrast, it includes more older and established companies as they could react differently during booms [51, 28, 68]. Those market datasets were pulled from Investing.com, which is a popular and trusted financial platform that provides access to historical data.

In addition, we used IPO data from Professor Jay Ritter at the University of Florida (link: https://site.warrington.ufl.edu/ritter/ipo-data/), who provides historical data of IPO markets. This data includes monthly number of IPOs and the average first-day return from 1994 to 2024, including both traditional IPOs as well as venture capital backed IPOs. In previous research, it has been suggested that we can identify over-hype by investors and rapid commercialization that are indicators of bubbles [30].

## 2.2 Analysis Methods

The analysis was approached differently for the scientific publication data and the market/IPO data, given how different they were in nature. We list those different analysis methods below.

### 2.2.1 Social Citation Network Analysis

With the scraped publication dataset of the two eras, we wanted to analyze the structure of research communities over time as a network, with the structure representing the scientific ecosystem. This would allow us to assess collaboration among scientists and disciplines within communities, diffusion of ideas, and development of sub-communities when comparing the dot-com era to the AI era for any similar patterns [29]. To do so, we created yearly citation social networks, using the Python packages igraph and NetworkX, where the authors are nodes and the edges are the relationships between two authors based on citations. An example of this would be Figure 1, which is a yearly citation network for authors who published in 1994 during the dot-com era. Direct relationships, of course, are created when the citing papers directly cite another author, while indirect relationships occur when an author cites a paper that cited another author without directly cited them (i.e. there is a “step” in between) [4]. By doing so, we were able to calculate three specific Leiden network characteristics for each year: network density, average path length, and collaboration counts. Density indicates how tightly a community is connected to one another by taking into consideration the number of connections present and the highest number of possible connections. The average path length indicates the number of steps, in average, it takes for knowledge to travel from one author to another. The collaboration count is used to see how often researchers are working with one another on projects, based on joint publications [34, 62].

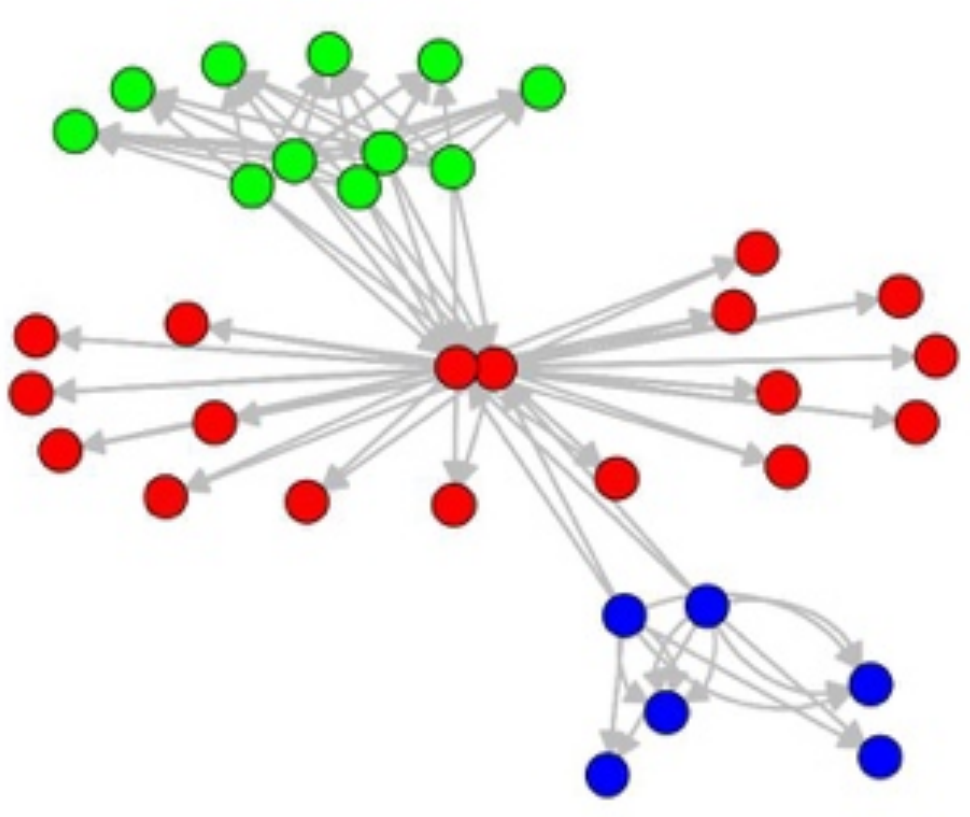

**Fig 1.** Author citation social network (1994)

Additionally, from the yearly citation networks, we collected information on clusters. Clusters are identified by the Leiden algorithm, which groups nodes that are highly connected via edges with one another and have fewer connecting edges to other clusters. Therefore, in our context, clusters are essentially a group of authors who often and repeatedly cite one another. This would mean that these authors are closely working with one another on common ideas and problems, which can be seen as the formation of specialized fields of study (sub-fields) within a broad field as emphasized by Newman (2006) in their study of social networks. The size of a cluster is the number of authors within a cluster. A large cluster would indicate that it is very popular and widespread with many authors actively contributing. However, a small cluster would indicate that it is a more of an upcoming and specialized area of study [69]. In order to further investigate the distribution of research activity, we utilized the Gini coefficient (calculated across all clusters within the era), which is a measure which is commonly used to explore inequality in economics such as GDP or income within a certain group of individuals [5]. The Gini coefficient here provides us with information regarding the fragmentation and concentration of these networks over time, based on the amount of variability (i.e. “inequality”) in cluster sizes.

### 2.2.2 Graph Convolutional Networks

Social networks provide us with information regarding the relationships of authors and their interactions through their work (scientific publications). However, in order to compare the relationships and interactions of authors between the two eras, we further utilized GCN models for a more in-depth analysis [75]. This model can allow us to explore whether the ***behavior*** of scientific communities during the dot-com era resembled that during the AI era (i.e. compare similarity of network structures). To do so, we used the through the NetworkX package in Python to calculated the betweenness centrality (BC), which indicates how frequently an author connects different groups of authors in a scientific community. Subsequently was calculated the "betweenness decay" (BD), which indicates how recently the author was a bridge between two communities by taking the betweenness centrality for each year and then dividing it by the reciprocal of $Year_B$ (bridge year) minus the max year of the period plus one (e.g. 2025 for the AI period). This is shown in Equation 1 below. The betweenness decay gives a greater weight to authors who were more recently influential [11, 74].

$$BD = BC * \frac{1}{Year_{max} - Year_B} \quad (1)$$

We are using the above two network characteristics because together they capture the relevance and role of authors in the networks who form a bridge between scientific communities that helps the flow of knowledge across fields. Using these features in the GCN modeling can aid in assessing the patterns of influence in the two eras for any possible similarities between how scientific communities function and communicate.

Two citation networks were built for each era where each author node has a betweenness centrality and decayed betweenness score, which was then passed to the GCN for modeling. For authors in the dot-com citation network, the top 1% of authors based on decayed betweenness scores were labeled as "influential" while the rest are labeled as "not influential" to create a labeled dataset for training a binary classifier, whereas for the AI network there were no labels since it is the "unseen" dataset to be predicted later. The author nodes of the dot-com citation network were split where 80% is for training and 20% is for testing. Since there was heavy class imbalance where 99% of nodes in the dot-com citation network are "not influential" but only 1% are "influential", the training set was oversampled to balance the data to ensure that the models perform well while the test set was left unbalanced to evaluate the model's ability to perform on realistic data. We performed hyperparameter tuning to ensure that the GCN model is performing maximally, then used AUC (area under the curve) to evaluate performance on the test set. The model that has the highest AUC score is then re-trained all of the dot-com citation network nodes, and finally applied on the AI citation network authors. Afterwards, each author node of the AI citation network received a predicted label (influential, not influential) and confidence score for the predicted label.

### 2.2.3 LSTM and KNN Phase Classifier

In order to explore whether the AI financial markets (CBOE, S&P 500, NASDAQ composite, NYSE, and IPO) are following a similar course to the financial markets during the dot-com bubble (CBOE, S&P 500, NASDAQ composite, NYSE, and IPO), we created models based on KNN and LSTM to perform phase classification and anomaly detection based on other approaches in the scientific literature [36].

The phase classification was trained on dot-com financial data that is labeled based on three phases. These are pre-bubble (1994 – 1997), build-up/hype (1998-1999), and burst (2000 – 2001). Then 10 month sliding windows with a 90% overlap of the market data were used as feature data points, with the phase label as the target variable. The missing values in the data were imputed with 0 and the features scaled using MinMaxScalar. This is because since we are dealing with time series and limited data, we don't want to drop any rows. The dot-com financial data is split 80% for training and 20% for testing for KNN modeling with 5-fold cross validation and evaluated through precision, recall, and f1-score. Then, the final model was trained on all of the dotcom data and applied to the AI period data to evaluate its predictions on that dataset.

The anomaly detection utilized an LSTM autoencoder, also based on 10 month sliding windows in the same manner with the financial data is split in the same way as well. Since this threshold captures 95% of expected behavior, then exceeding it means that the behavior it is observing is significantly different from what the model learned previously, which could possibly indicate that there is a new pattern or phase. Then the model was tuned and evaluated through validation loss (MSE). Afterwards, the final model is trained on all of the dotcom data, and likewise applied to the AI period to evaluate its predictions there.

For evaluating the predictions on the AI era financial market data, we sequenced it into 10 month sliding windows then predicted phases for the end of each window. Meanwhile, the LSTM autoencoder attempted to reproduce the AI sliding windows via an encoder/decoder structure, which then allowed us to calculate an MSE score. This MSE score is compared with an "anomaly threshold", which is defined as the mean MSE plus 2 standard deviations [17]. sequences that cross this threshold indicate new patterns that have not been seen previously during that era, which could indicate impending phase shifts.

#### 2.2.4 AR-X and GARCH

To evaluate the interaction of the scientific publication data and the market/IPO data, we used two models: AR-X and GARCH. Those two models were used to explore whether scientific network characteristics can possibly explain market changes and, more specifically, the underlying volatility in the financial markets (CBOE, S&P 500, NASDAQ composite, and NYSE), as suggested in previous literature [59]. The AR-X explored the relationship between historical data (both financial returns and research activity) with future financial returns. Meanwhile, GARCH was used to identify temporal changes in market volatility over time. To do this, the monthly returns of the financial markets were logged and then used as the target variable. Then, this is combined with feature data from the scientific network characteristics described in Section 2.1.1, such as collaboration counts, average path length, and density. Taking into account the potential lag of scientific innovations impacting the markets, we included a lag time of 12 months between the feature data and target data. We chose 12 months because we wanted to avoid sensitivity to short-term changes in the market. By using this modeling, we can identify whether research activity from publication networks is linked to future behavior of financial markets.

## 3. Results

### 3.1 Social Citation Network Findings

To examine the evolution of scientific publishing behavior of authors over time as it relates to financial market cycles, we analyzed yearly citation social networks during the two eras in question (dot-com era and the AI era), the results of which we present in separate graphs below for comparison

For the dot-com era, Figure 2 consists of plots that visually display the quantitative changes of clusters in scientific publication networks. These plots include a trend line, which helps to identify the overall pattern and/or direction of the data, and a Pearson correlation coefficient (r-value), which measure the linear correlation such as strength and direction of two quantitative variables. Figure 2A indicates that there is an moderate decrease in the average cluster size ($r = -0.59$), whereas the number of clusters (see Figure 2B) has a weak upwards trend due to the low-to-moderate r-value (value = 0.44). This indicates that as time passed, the average number of authors in these clusters steadily decreased and there was a general increase in the number of clusters, though the latter was somewhat inconsistent year-to-year. Furthermore, Figure 2C shows that the Gini coefficient of cluster sizes decreased and that clusters became more "equal" sized with each other, which, in our context, could indicate that research efforts became more equally distributed (perhaps due to the fragmentation into smaller groups). To further probe into the underlying patterns of the dot-com networks, we created Figure 3, which shows a graph of quantitative measures such as density (network density), path length (average path length), and collaborations (number of collaborations) that are plotted against each other over time. The y-axis in Figure 3 is min-max scaled (0 to 1) for all those metrics. The corresponding r-squared values for all of those measures in Figure 3 are shown in Table 1. As seen in the trend lines in the figure, there was a general increase in density and decrease of path length. This would mean that there was an increase in author-author connections, hence, making the network denser with fewer steps needed for knowledge and findings to spread from one author to another. However, the trend line for collaborations indicates that there was decrease in collaborations among individual authors, though with a fairly low r-value of -0.300 (Table 1) indicating inconsistency in that decline year-to year. To summarize, in the dot-com era, the network over time appears to have fragmented into smaller and more even-sized clusters or sub-fields, even while author citation networks became denser and more interconnected with one another. At the same time, the decline in direct collaborations between authors, though inconsistent, may reflect that research partnerships were becoming brief and relationships between authors were becoming more "indirect", which has been previously suggested in the academic literature [40].

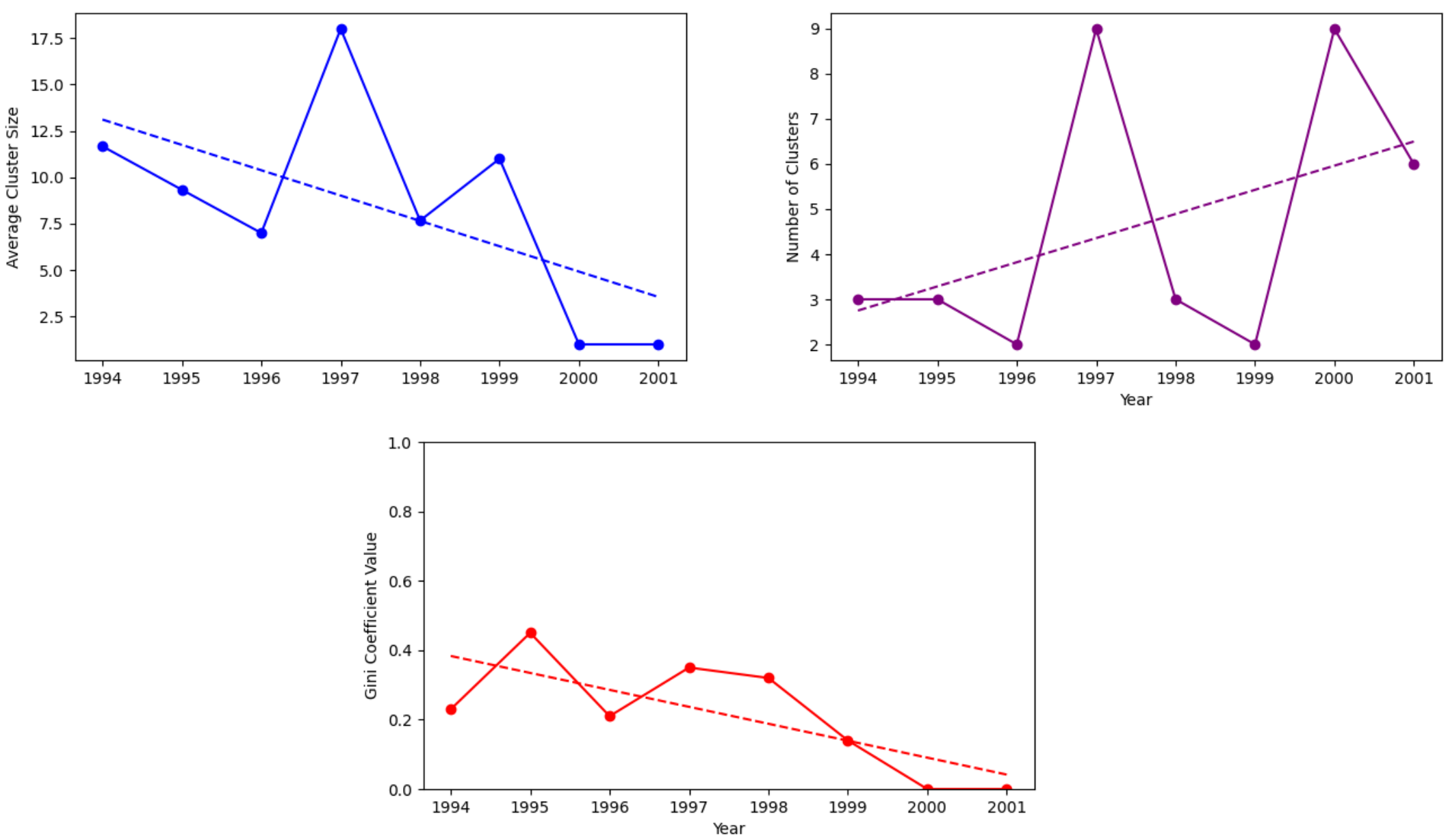

**Fig 2. (Top Left) A** Average cluster size during the dot-com era with r-value = -0.59, (Top Right) **B** Number of clusters during the dot-com era with r-value = 0.44, **(Bottom) C** Gini coefficient during the dot-com era with r-value= -0.74.

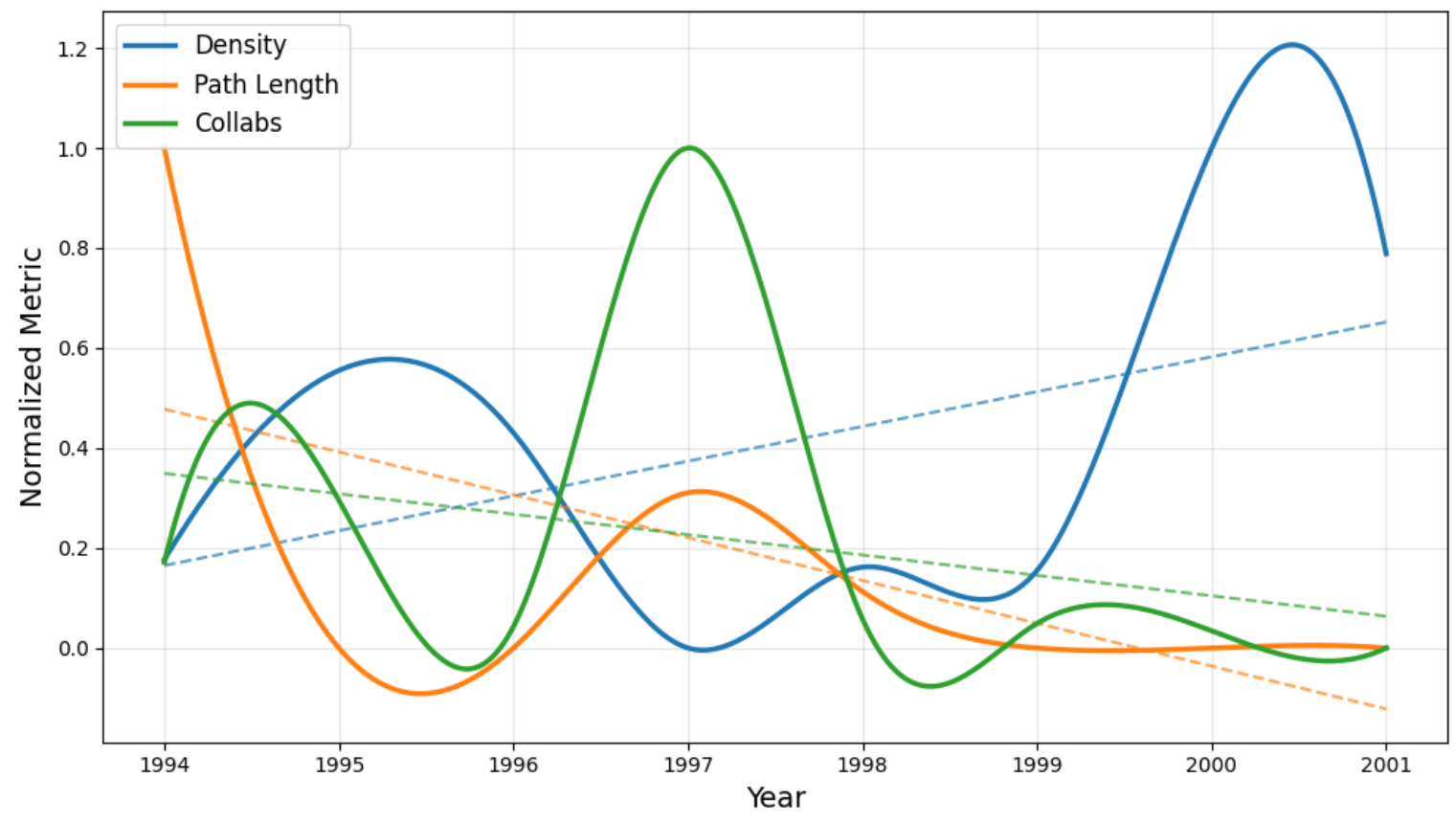

**Fig 3.** Network characteristics plot for the dot-com citation social networks.

**Table 1.** R-squared summary of network in the dot-com era.

| Network Characteristic | Pearson Correlation Coefficient (r) Value |
|---|---|
| Density | 0.490 |
| Path Length | -0.600 |
| Collabs | -0.300 |

For the AI era, Figure 4 shows the changes in clusters while Figure 5 shows the changes in network characteristics over time, which can be compared to Figure 2 and 3 above, respectively, for the dot-com era. Table 2 contains the corresponding r-squared values for Figure 5, which can be compared to Table 1 for the dot-com era. For Figure 4, there is a strong upwards trend in the average size of clusters with a high r-value of 0.93 (Fig. 4A) and in the number of clusters with an r-value of 0.84 (Fig. 4B). In other words, there is an increasing number of authors joining clusters or sub-fields over time, while at the same time the number of clusters is increasing. On the other hand, though the trend for the Gini coefficient indicates a moderate increase (r-value = 0.53), the value across all years in the AI era is consistently above the highest Gini value for the dot-com era in 1995 (roughly 0.4), even in the initial year 2017. That suggests greater inequality in sizes of clusters, where some are much larger or smaller compared to others, has been a consistent theme during the AI era. Moreover, the cluster size variation is not changing over time as in the dot-com era, but rather gradually increasing within a narrow range. To summarize, the r- values are much higher for the AI era than the dot-com era, with quite different trends in cluster changes over time indicating that publishing activity also differed.

When we look at Figure 5, there is a strong increase in collaborations (r-value = 0.881). Meanwhile, density decreases and path length increases over time, which suggest that authors are more spread out from one another and it takes more "steps" for knowledge to spread across authors throughout the network. Compared to Figure 3, we can say there are stark differences between the AI era (Figure 5) and dot-com era (Figure 3) in terms of the network characteristics of publishing activity during those eras. Overall, the evidence suggests that in the post-2000 AI period , there appear to be more authors who are working with one another in larger variably-sized clusters or sub-fields within AI (e.g. more collaborations). However, at the same time the scientific publication network is becoming more segmented and spread out (e.g. longer path lengths and low density), which can cause ideas and knowledge to take longer to spread, similar to the network patterns found in echo chambers on Twitter/X [16]. Those trends may be leading to scientific papers becoming less disruptive/impactful over time during the AI era, as other research has argued [44].

In summary, the patterns of publishing we observe in the AI era (Figure 4 and 5) do not resemble trends from the dot-com era (Figures 2 and 3), hence, suggesting that the indicators from the dot-com era may not be relevant in forecasting an AI bubble or its potential burst. However, it may also be due to changes in the scientific publishing ecosystem between the 1990's and the 2020's, which we discuss further in the Discussion section below.

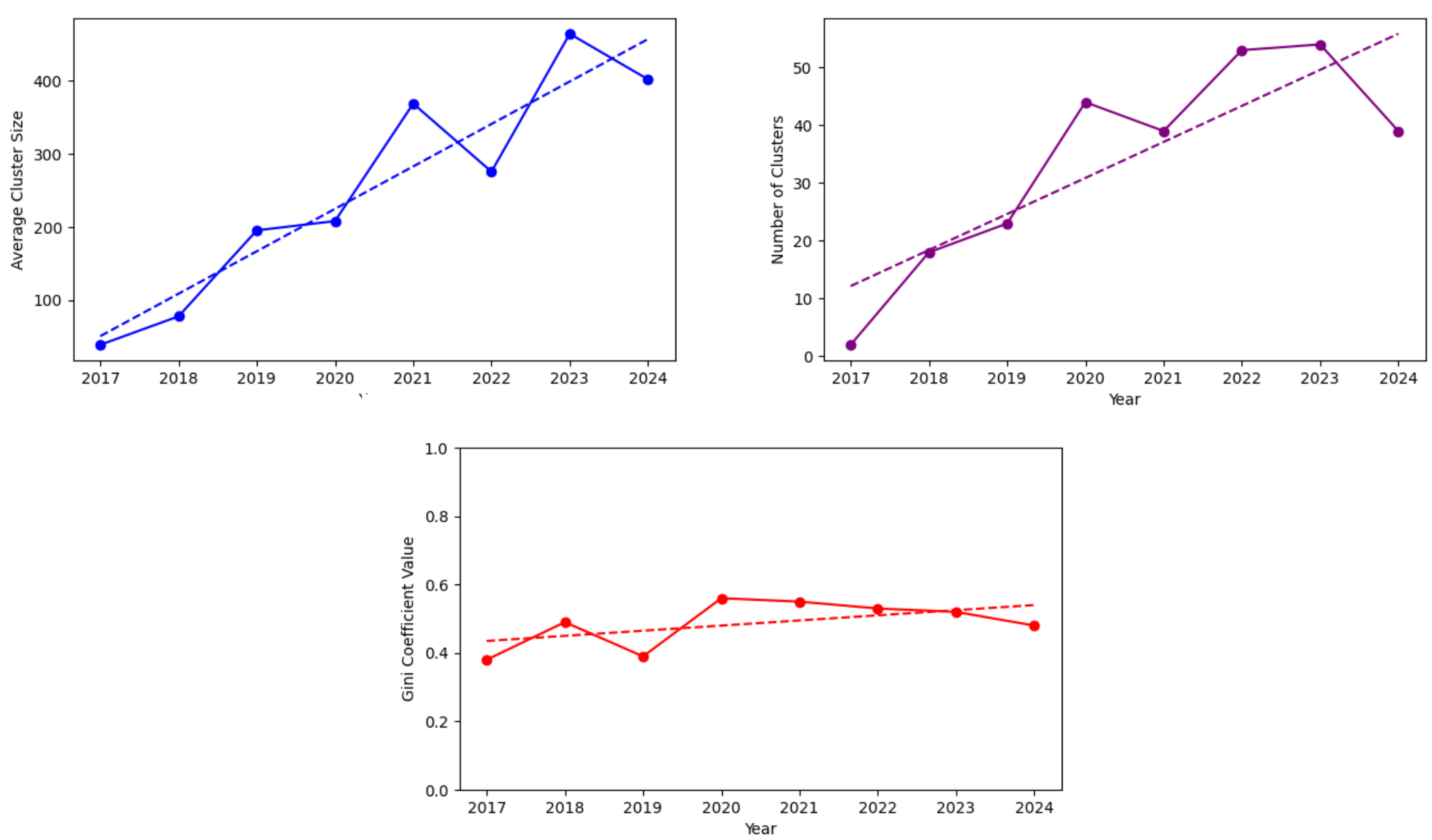

**Fig 4. (Top Left) A** Average cluster size during the AI era with r-value = 0.93, (Top Right) **B** Number of clusters during the AI era with r-value = 0.84, **(Bottom) C** Gini coefficient during the AI era with r-value = 0.53.

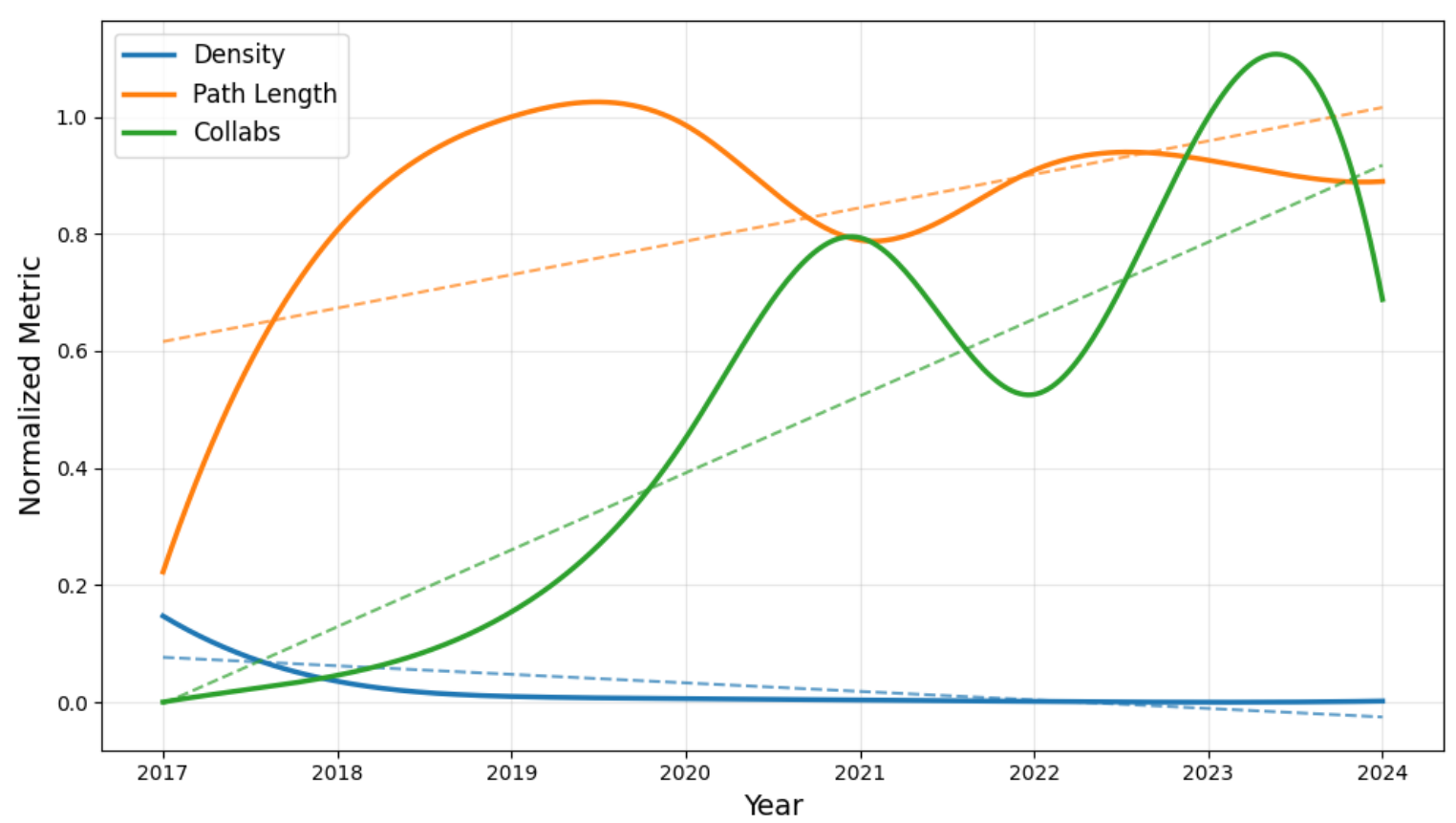

**Fig 5.** Network characteristics plot for the AI citation social networks.

**Table 2.** R-squared summary of network in the AI era.

| Network Characteristic | Pearson Correlation Coefficient (r) Value |
|---|---|
| Density | -0.707 |
| Path Length | 0.557 |
| Collabs | 0.881 |

### 3.2 Graph Convolutional Networks Findings

Through the yearly citation social networks in Section 3.1, we were able to compare and contrast the eras via general network characteristics such as density, path length, and collaborations. However, in order to understand and compare the more subtle behaviors and influence of scientists in those networks, we used a GCN to conduct a more in-depth analysis of network structures [75]. Though that, we discovered that the average influence of all the scientists in the dot-com era differed quite a bit from those in the AI era, through mean author embeddings. These embeddings capture the average influence of each author on the rest of the network, which can be thought of as similar to the "impact factor" of their scientific research and publications. When the mean author embeddings between the dot-com and AI eras are compared based on the cosine similarity of the mean embeddings (i.e. average behavior of all authors combined), we calculated a value very close to 0 (value = 3.09e-07). In other words, that cosine value means that the influence patterns of authors between the two eras are completely different and reflect distinct structural patterns from one another.

However, we do *emphasize* that there were a number of scientists (1,964 out of 285,773 total scientists) in the AI era network whose average influence values very closely resembled the average influence values of highly-influential scientists in the dot-com network. This could mean that, although the AI citation social network appears to be evolving differently compared to the dot-com citation social network and has not yet shown similar dense inter-connectedness of that era, **there are a subset of AI scientists who are behaving similarly to influential dot-com era scientists acting as strong bridges between sub-fields within the overall scientific community.** This finding may serve as an early indicator for potential hype, but it is not definitive and would need further study focused on how those influential network nodes affect the rest of the network over time.

### 3.3 LSTM and KNN Phase Classifier Findings

When we applied the LSTM and KNN phase classifier model trained on the CBOE index and IPO data during dot-com era data to the AI era, it did not seem to correspond to any obvious growing AI bubble. Figure 6 visualizes those results, where yellow areas represent pre-dot-com bubble phase, orange areas represent hype/build-up of dot-com bubble phase, and red areas represents the burst of the dot-com bubble phase. However, there was of course no AI bubble burst during 2023 (as of this writing in 2025), despite what one might expect from the model Figure 6.

Investigating the results in Figure 6 further, we found that from 2020 to 2021 the forecasted risk passes the anomaly threshold, which indicates that the model is coming across behavior in the AI era financial data it has not seen before in the dot-com era financial data. At the same time, the model classified this time period to have a hype/build-up phase environment similar to the dot-com era. This means that the investor sentiment and IPO activity seen in this period of time is similar to what was seen during the dot-com era, but the anomaly indicator suggests there is some behavior that is a bit different and previously seen in the dot-com financial data. In a way, this can be seen as a risk because such new anomalous behavior is obviously not predictable via a model and thus can cause consequences that we are not prepared for. In 2023, we can further see that the model classified this period as the bubble burst phase while the forecasted risk is still relatively low. This probably indicates that this behavior has been seen before during the dot-com era, but we obviously did not have an AI "bubble burst" during 2023 [42]. One possible explanation is **that could be because the financial markets have changed and evolved in the 2020's versus the 1990's [31].** In the financial data, we are not seeing any real-world events related to the classified phases in Figure 6, but this could mean that the potential AI bubble is taking shape in a different way or form compared to what we observed during the dot-com era and using such method does not hold predictive value in the AI era [19]. In contrast, it could also indicate there is no AI bubble forming. Both explanations are feasible, but based on our data here we can't confirm either one conclusively.

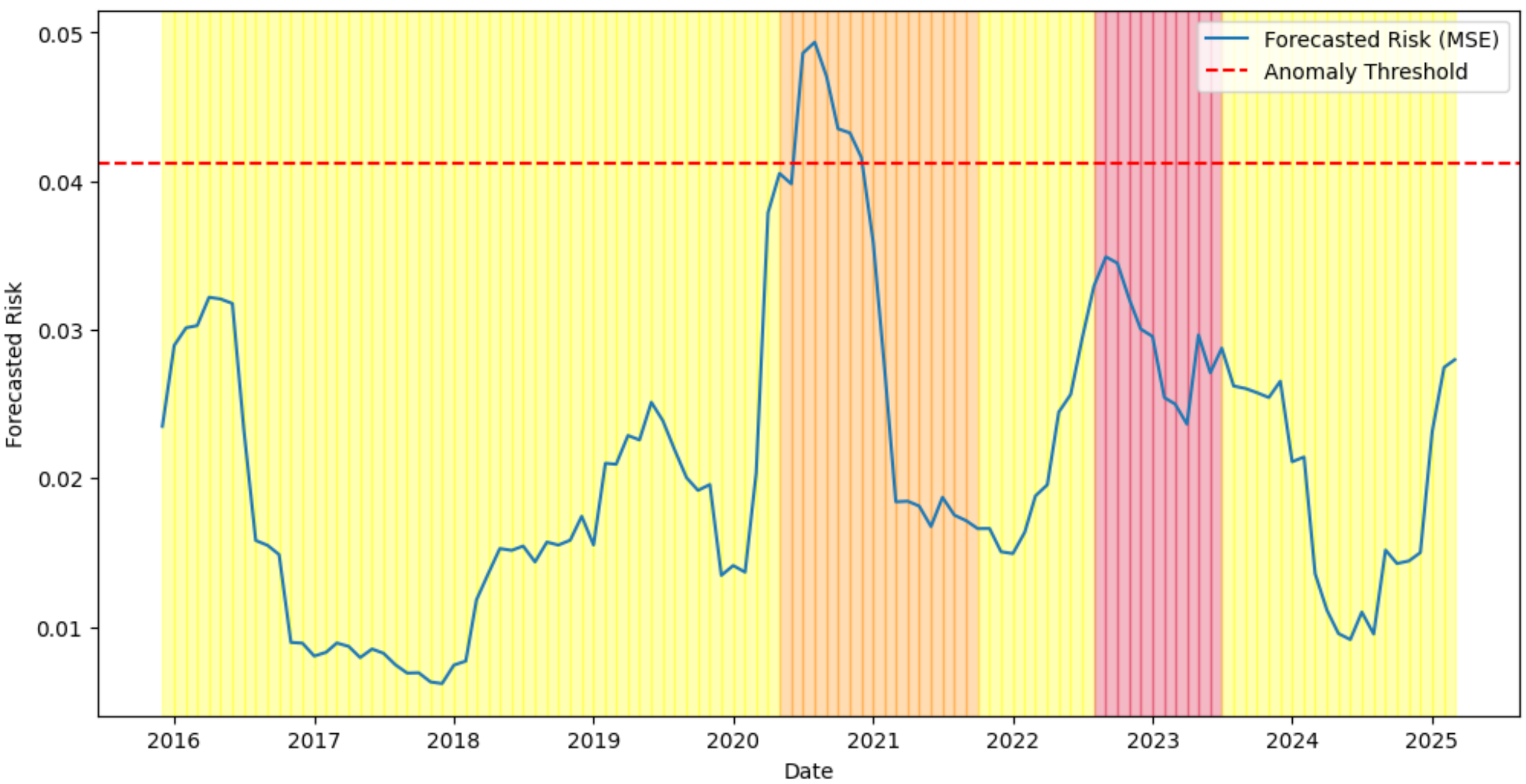

**Fig 6.** Plot produced by the LSTM + KNN phase classifying model on CBOE Volatility Index (VIX) and selected IPO features.

We should note that when we applied the LSTM and KNN phase classifier model trained on the NYSE, NASDAQ, and S&P 500 indices during the dot-com, it was not able to appropriately classify different phases with any coherency. The failure of the model could be happening because it is unable to capture the subtle changes in those indices used, which is a problem reported in previous literature [67]. However, the model was able to differentiate various phases for CBOE Index and IPO features, as shown above. Thus, that could be because CBOE Index and the IPO features are more sensitive and change more quickly to various investor sentiments like fear, excitement, and so forth, in contrast to those other stock indices [18, 49].

### 3.4 AR-X and GARCH Findings

We wanted to look further into the market volatility of financial markets in relation to citation social networks to possibly find and compare/contrast any patterns between the dot-com bubble and the potential "AI bubble" for speculative investing behavior. We did this using two companion models: AR-X and GARCH. The AR-X part of the model predicts the financial return using lagged time (12 months) to measure the impact of the selected citation network characteristics (collaborations, path length, and density) of research activity with *future* financial returns. The GARCH portion is used to track volatility in the indices. More specifically, the GARCH method is used to measure the impact of volatility (sensitivity and reactivity) of the indices over time as it relates to future returns [48]. From the GARCH model, we calculated the *alpha* value, which measures the impact of sudden changes in future volatility, and *beta* value, which measures the duration of impact for such volatility-related events.

We collated all of our findings and organized them in Table 3 and Table 4. For the dot-com era, as seen in Table 3, CBOE results showed that financial returns were influenced by scientific collaboration levels (value = 0.362, p-value = 0.01). In contrast, path length had a negative relationship, but that was not significant (value = 0.285, p-value = 0.058). Likewise, the density value was also not significantly related. For the GARCH values, the alpha value was high at 0.856 and the beta relatively small at 0.144, which appears to indicate that the shocks in the CBOE index during the dot-com era had a strong impact on volatility but did not have a lasting impact. However, for NYSE, NASDAQ, and S&P 500, none of the coefficients were significant as their p-values are greater than 0.05. This indicates that the changes in the network have little to no influence on the price movements.

When we performed the same modeling on AI era data, as seen in Table 4, only NYSE, NASDAQ, and S&P 500 showed significant findings while CBOE did not. NYSE, NASDAQ, and S&P 500 showed significant positive impacts with the changes in the network characteristics. For instance, as collaborations increased, it also increased the volatility of the indices: NYSE (value = 0.062, p-value = 0.02), NASDAQ (value = 0.085, p-value = 0.014), and S&P 500 (value = 0.066, p-value = 0.017). The alpha was higher and the beta values lower compared to what we saw in the dot-com era, which means that there is less impact of sudden changes in the market overall (i.e. investors seem more tolerant of volatility), though with slightly less lasting effect. This is different from what we saw in the dot-com era where volatility was highly dependent on collaborations and path length. Rather, the indices seem to be reflecting changes in the citation network characteristics.

From these findings**, it seems like in the AI era the stock market indices such as NYSE, NASDAQ, and S&P 500 are being more impacted by the publishing behavior of scientists working in the field as compared to the dot-com era.** Additionally, even though the indices are becoming more volatile in the AI era there is less impact on future returns, which could reflect the overall consensus at the moment of investors being excited by the hype of AI [2]. This is different from what we saw in the dot-com bubble where scientific behavior had less impact on volatility, and thus the patterns in dot-com era does not effectively translate to the AI era [71].

**Table 3.** AR-X and GARCH findings for the Dot-Com Era, significance levels *<0.05, **<0.01, ***<0.001

| Index | Collaborations Coefficient (p-value) | Path Length Coefficient (p-value) | Density Coefficient (p-value) | Alpha | Beta |
|---|---|---|---|---|---|
| **CBOE** | 0.362 (0.01*) | -0.285 (0.058) | -0.088 (0.51) | 0.856 | 0.144 |
| **NYSE** | 0.013 (0.765) | -0.001 (0.982) | 0.002 (0.949) | 0 | 0.935 |
| **NASDAQ** | 0.044 (0.676) | -0.120 (0.372) | -0.183 (0.121) | 0 | 1 |
| **S&P 500** | 0.012 (0.823) | -0.014 (0.835) | -0.036 (0.369) | 0 | 0.951 |

**Table 4.** AR-X and GARCH findings for the AI Era, significance levels *<0.05, **<0.01, ***<0.001

| Index | Collaborations Coefficient (p-value) | Path Length Coefficient (p-value) | Density Coefficient (p-value) | Alpha | Beta |
|---|---|---|---|---|---|
| **CBOE** | 0.011 (0.935) | -0.234 (0.688) | -0.411 (0.903) | 0.218 | 0.135 |
| **NYSE** | 0.062 (0.02*) | 0.362 (0.0005***) | 1.81 (0.004**) | 0.479 | 0.000 |
| **NASDAQ** | 0.085 (0.014*) | 0.468 (0.004**) | 2.53 (0.007**) | 0.251 | 0.388 |
| **S&P 500** | 0.066 (0.017*) | 0.346 (0.003**) | 1.78 (0.008**) | 0.340 | 0.204 |

# 4. Discussion

## 4.1 Summary

In summary, we utilized temporal SNAs to detect if there was any relationship between the publication citation networks of scientists and financial bubbles during two eras: dot-com era of the late 1990's and the AI era from 2017-2024. The dot-com era of course ended in a massive bubble burst in 2001. Our approach consisted of analyzing scientific publishing patterns and financial data (such as stock market indices and IPO data) from the dot-com era to assess – when applied to the AI era - whether it is possible to identify early-warning indicators of a potential bubble formation/burst from the research activity and financial data in the AI era. From the findings, we discovered indications that it does not appear we cannot confidently predict or identify an AI bubble through such an approach. To support our claim, we present a range of evidence that stems from multiple types of analysis on the datasets in both eras.

Such evidence includes the stark differences in the patterns of publishing in the yearly citation social networks during the two eras. We observed that in the AI era compared to the dot-com era, the indicators from the citation network characteristics, such as density, path length, and collaborations, were completely different and thus not relevant in forecasting the presence and potential burst of any AI bubble (Section 3.1). We also further evaluated the behavior of scientific communities and found that despite the completely different overall influence patterns of authors between the two eras (based on citation levels), there does appear to be a subset of AI scientists who are behaving similarly to scientists from the dot-com era (Section 3.2). This finding has the potential to be useful signal for the emergence of hype, but it is inconclusive at the moment and needs further study, which we discuss later in the future work section. Moreover, an analysis of the financial markets of the two eras indicate that either the potential AI bubble is taking on a form we have yet to witness historically in the financial markets or, alternatively, that there is no AI bubble forming. However, both those possibilities require further study to confidently confirm either one (Section 3.3). Finally, we looked at market volatility as a measure of hype, and whether that can be predicted using identification publication activity levels using AR-X and GARCH models (Section 3.4). That analysis showed that stock market indices such as NYSE, NASDAQ, and S&P 500 are becoming more volatile over time and there appears to be a significant correlation to publishing behavior of scientists in the AI era. That correlation was not seen when analyzing the dot-com era data, which suggests that the patterns in the dot-com era do not appear to carry over to the AI era and that the trends observed in the dot-com era have limited applicability to the current market environment.

### 4.2 Interpretations

The overall picture from our findings is one of ambiguity. While, the findings did not allow us to definitively predict an AI bubble or its potential burst, there are two possible explanations as to why. One is that there is no AI bubble forming in the near future. The other is that, alternatively, the AI bubble is taking on a form that is structurally different to what we have seen before during the dot-com era. These possibilities have the potential to stem from changes in either scientific publishing behavior or investor behavior, or both perhaps. We consider both explanations in detail below.

There has been a rapid growth in the overall number of scientific publications in recent years [10, 15, 35, 44]. This boom in publishing not only presents many opportunities, but challenges as well. There is heightened attention towards AI and the benefits it brings to humanity [43]. On the other hand, there is a risk that the increased publishing volume will mask the usual indicators of scientific advancement and/or scientific impact. This is because as the rate of publication increases, it has the potential to alter the structure of scientific research networks by supercharging things like impact factor metrics (i.e. more citations just due to more publications overall) compared to historical levels. That can undermine the utility of such metrics as a measure of scientific quality, if they become artificially inflated due purely to greater publication quantity. The volume shift also further complicates the identification of early-warning signals using traditional network analysis metrics such as density, path length, and collaborations, as those may not be as useful as predictors. Therefore, the explosion publication volume coinciding with the rapid growth of AI has the potential to introduce noise into the analysis that can overshadow patterns related to typical hype-cycles, which could explain the differences in network characteristics between the dot-com and AI eras (Section 3.1).

Meanwhile, there is also the possibility that investor behavior has changed since the dot-com era. In late 1990s and early 2000s, investors were often chasing momentum and the availability of real-time in-depth stock information was limited given the early stage of the internet at the time [39]. That drove highly inflated stock prices in the dot-com era through pure excitement based on limited information, which raised the valuations of stocks even more beyond what technical fundamentals might have suggested. Later when progress and growth slowed in the dot-com companies, investors essentially wiped-out inflated valuations by rushing to sell their shares rapidly. It thus could be argued that, in the dot-com era, investors rushed into the market without fully considering the risks of the equities and assets they were investing in. In contrast, in the AI era, it is possible that investors are more “tolerant” per say to the volatility in the markets due to more widely available information via (ironically) the internet. In other words, compared to the dot-com bubble where investors would highly react to sudden short-term market changes, in the AI era, investors do not appear to be reacting with as much sudden excitement and concern. Rather, they seem willing to endure sudden drops in the market and ride the ”hype”. That is a possible explanation.

However, we do want to emphasize here that both explanations above are simply assumptions attempting to explain patterns in the results here, and that further research is needed to evaluate them empirically.

### 4.3 Scientific “Economy” and the Breakdown of Peer Review

Scientific publishing in 2025 is quite different from 1995. As we mentioned previously, there has been a rapid growth in the overall number of scientific publications in recent years, along with the growth of practices such as open-access publication [10, 15, 35, 44]. At the same time, there is growing awareness of an overall "crisis" in the peer review process for scientific publications, with a lack of available qualified peer reviewers. To put it bluntly, if we assume that any random scientific paper has a 25% chance of getting accepted, and all publications need at least 2 reviewers, then any scientist should technically be reviewing at least 2 papers for each paper they publish every year, though some published estimates have stated as high as 3-4 [50]. So, if college professor X published 5 papers last year, then they should have peer reviewed 10-20 papers. However, there is no current incentive for scientists to do so, outside of some personal honor code. Universities typically do not factor peer reviewing into decisions on promotion or tenure, for that matter [13, 33]. Nor does peer reviewing contribute to impact factor metrics for individual scientists, which leads to some perverse incentives for scientists to publish a lot but review very little.

Technically speaking, scientific publication has always operated as an *informal economy*, where scientists submit papers to be peer-reviewed by other scientists, with the general understanding that they return the favor-in-kind and careers on-the-line [26, 45, 58]. Yet such informal economies are always at risk of falling prey to the "tragedy of the commons" [7]. **As such, perhaps the time has come to *formalize* that scientific economy for publications**, to counter the peer review crisis and perverse incentives. We propose that could be created rather simply by instituting "tokens" (e.g. via blockchain), which scientists would receive after conducting a peer review. This is similar to other suggestions in recent years [37, 56]. Those tokens could then be used as credits when they submit their own papers for publication, with each submission requiring some number of credits. To make it fairer, early-stage researchers could receive a pool of tokens to start with, and perhaps tokens could be partially refunded if a paper was rejected after peer review. Furthermore, peer reviewers could be rated on the quality of their peer review by the editors of the journal/conference, which might affect how many tokens they receive and thus ensure they make a reasonable effort. In essence, this would create a more formal scientific economy that is less likely to be gamed and more resistant to behaviors engendered through perverse incentives.

The difficult part of the above would be the administration of such a system. In real economies, that is of course handled by governing bodies at a national level. However, science is an international effort at this stage, with many different actors. Thus, the challenge is to ensure the integrity of such a formal science economy, and whether the token system would be managed by one large journal publisher, or some group of publishers, or perhaps a non-profit organization on behalf of scientists. Also, there is an open question of whether all publishers (journals and/or conferences) would be included, or whether a publication venue's participation in the token system would be restricted somehow. The problem is if it is not restricted, that may lead to abuse whereas small groups of scientists could setup "sham" conferences or journals to inappropriately obtain tokens. All that said, there is already some precedent for what we are suggesting, in the form of international bodies like Science Citation Index (SCI) or Web-of-Science (WOS) or ORC-ID that could perhaps be used to restrict access to the token system based on some sort of quality-control, in order to prevent any abuse.

Regardless, we think the above concept is something that needs to be seriously discussed across the scientific community, as this paper and other recent papers on the topic have shown we are likely reaching a critical point in the history of science, and that scientific practices we have traditionally used in the past are no longer suitable for current conditions [44]. Addressing this concern has the potential to curb the trend of excessive publishing so that there is a shift from volume to more rigorous and high-quality work in the scientific community, and that citation networks more accurately reflect actual influence of individual papers and/or scientists. Such guardrails would most likely help with reducing the noise in early-detection models of scientific hype like the ones used in this paper, therefore, potentially mitigating risks involved with financial bubbles in the future.

### 4.4 Limitations

This study has several potential limitations. For example, when scraping from the OpenAlex publication database, there exists a 200-citation cap set by OpenAlex that restricted the dataset size. Of course, the vast majority of papers do no reach that cap (less than 1% exceed 100 citations) [65], but nevertheless it is a technical limitation of using the OpenAlex database. While this may have not had a large impact on the analysis, it has the potential to introduce some bias to publications that are highly influential and exceed the threshold of 200-citations. Another limitation is that the AI era dataset ends in 2024, but the speculated "AI boom" is still continuing to this day and thus more recent AI era papers may still be accumulating citations [23]. Although the two eras were matched in length in our analysis in terms of years, the context is different as the selected window for the AI era may be too short and/or recent to observe such bubble behaviors. Another limitation is that financial indices are highly sensitive and constantly changing for various reasons that have nothing to do with scientific activity, such as geopolitical changes and economic announcements [32]. Such complexity results in models having a difficult time in identifying long-term trends and forecasting fluctuations, given the wide range of factors involved. Finally, citation networks may not always represent the flow of knowledge as it pertains to later creating actual products within the economy. Instead, it is possible that researchers cite one another for other reasons such as following certain conventions or even bias within the scientific community itself [73]. This would indicate that not every citation is necessarily a true representation of the flow of ideas.

### 4.5 Future Work

While our work provides a framework for the comparison of the two posited tech bubbles (dot-com and AI bubble), it has its limitations. Therefore, more research is needed on this area in order to open up future exploration to discover and validate early-warning signals and patterns of financial bubbles based on scientific activity.

Firstly, by incorporating other bibliometric data such as preprints and patent filings along with other financial data such as startup funding rounds, we could help capture more innovative research (sometimes from industry scientists, not academia) and thus reveal a bigger picture of the dynamics in the tech market. Furthermore, tracking the data for a longer period of time for the AI era could reveal more confident signals about the presence of the AI bubble and its potential burst, though we would need to figure out appropriate time scales and/or scaling for different window sizes. There could also be a potential system that could be monitor and track the patterns in real time, which could also be accessed by policy makers and/or the general public. Moreover, further network analysis can be performed utilizing other graph network characteristics, such as centralization and interdisciplinary metrics so that the shift of clusters over time can be tracked on a deeper basis.

Investor sentiment is also important to consider when dealing with financial market data, of course, and this research could be extended in that direction. Further investigation into other quantitative indicators along with more sentiment-based analysis of public media (i.e. financial newspapers, social media, and analyst reports) could explore the relationship between emotions and investment decisions and whether it has an impact on financial bubbles, potentially acting in concert with the effects of scientific publishing activity. Finally, comparing other financial bubbles, such as emerging ones in the cryptocurrency markets, could make for an interesting comparison to the proposed AI bubble, using the same analysis approach we used here.

## Acknowledgements

We would like to thank our many research collaborators for their thoughtful discussions on this topic prior to this publication, as well as some of our scientific colleagues who contributed to the initial idea behind this research. This research involved no funding, nor did we receive any financial compensation from any publicly-traded companies related to this work.